%% file: main.tex
\documentclass[10pt,twocolumn,letterpaper]{article}

\usepackage[pagenumbers]{cvpr} 

\input{preamble}

\definecolor{cvprblue}{rgb}{0.21,0.49,0.74}
\usepackage[pagebackref,breaklinks,colorlinks,citecolor=cvprblue]{hyperref}
\usepackage{mathtools}

\DeclarePairedDelimiter\floor{\lfloor}{\rfloor}

\definecolor{gold}{RGB}{255, 215, 0}
\definecolor{silver}{RGB}{192, 192, 192}
\definecolor{bronze}{RGB}{205, 127, 50}

\def\paperID{234} 
\def\confName{3DV\xspace}
\def\confYear{2025\xspace}

\title{FourieRF: Few-Shot NeRFs via Progressive Fourier Frequency Control}

\author{Diego Gomez, Bingchen Gong, Maks Ovsjanikov\\
LIX, École Polytechnique, IP Paris\\
Palaiseau, France\\
}

\begin{document}

\twocolumn[{%
\renewcommand\twocolumn[1][]{#1}%
\maketitle
\vspace{-2em}
\includegraphics[width=\linewidth,trim={0 0 -2em 0},clip]{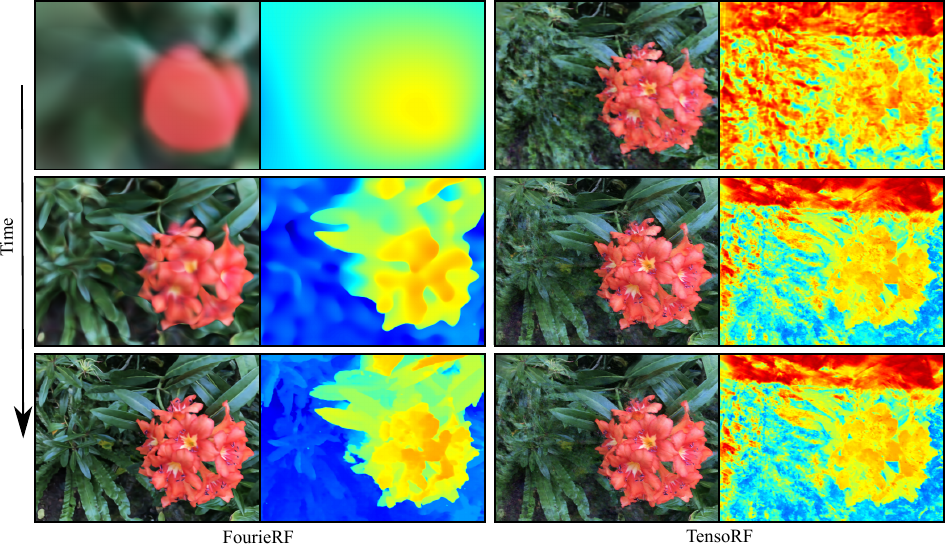}
\captionof{figure}{\textbf{FourieRF} serves as an effective and simple baseline for tackling the few-shot rendering problem. The vanilla approach often encounters high-frequency artifacts early in the optimization process. We introduce an explicit curriculum training procedure that gradually incorporates higher frequencies to mitigate this. This method ensures a stable training trajectory, eliminating major artifacts and enhancing overall rendering quality. \vspace{0.5em}}
\label{fig:teaser}
}]


\input{sec/0_abstract}
\input{sec/1_intro}
\input{sec/2_related_work}

\input{sec/3_motivation}
\input{sec/4_method}

\input{sec/5_experiments}
\input{sec/6_conclusion}

\newpage
{
    \small
    \bibliographystyle{ieeenat_fullname}
    \bibliography{main}
}
\input{sec/X_suppl}

\end{document}

%% file: preamble.tex
\usepackage{booktabs}
\usepackage{multirow}
\usepackage[dvipsnames]{xcolor}
\usepackage{colortbl}

\newcommand{\first}[1]{\cellcolor{red!25}#1}
\newcommand{\second}[1]{\cellcolor{orange!25}#1}
\newcommand{\third}[1]{\cellcolor{yellow!25}#1}

%% file: sec/0_abstract.tex
\begin{abstract}
\vspace{-1em}
We present a novel approach for few-shot NeRF estimation, aimed at avoiding local artifacts and capable of efficiently reconstructing real scenes. 
In contrast to previous methods that rely on pre-trained modules or various data-driven priors that only work well in specific scenarios, our method is fully generic and is based on controlling the frequency of the learned signal in the Fourier domain.
We observe that in NeRF learning methods,  high-frequency artifacts often show up early in the optimization process, and the network struggles to correct them due to the lack of dense supervision in few-shot cases. 
To counter this, we introduce an explicit curriculum training procedure, which progressively adds higher frequencies throughout optimization, thus favoring global, low-frequency signals initially, and only adding details later. We represent the radiance fields using a grid-based model and introduce an efficient approach to control the frequency band of the learned signal in the Fourier domain.
Therefore our method achieves faster reconstruction and better rendering quality than purely MLP-based methods.
We show that our approach is general and is capable of producing high-quality results on real scenes, at a fraction of the cost of competing methods. Our method opens the door to efficient and accurate scene acquisition in the few-shot NeRF setting. 
\vspace{-1em}
\end{abstract}

%% file: sec/1_intro.tex
\vspace{-1em}
\section{Introduction}
\label{sec:intro}


The introduction of Neural Radiance Fields (NeRFs)~\cite{mildenhall2020nerf} has marked a significant milestone in the realm of 3D scene generation from 2D images using neural networks.
NeRFs create continuous 3D scene representations by predicting color and density from various viewpoints, allowing them to synthesize photorealistic novel views from perspectives not included in the training data.
This breakthrough has revolutionized applications in novel view synthesis, 3D asset generation, and inverse rendering, enabling unprecedented accuracy and realism in these domains~\cite{mueller2022instant,Jin2023TensoIR,NeRFshop23}.

However, one major challenge with NeRFs is their need for a large number of input images to ensure accurate scene reconstruction~\cite{yu2021pixelnerf,yang2023freenerf}.
This limitation highlights the importance of the \textit{few-shot rendering problem}, which aims to perform novel view synthesis with only a limited set of input views. Advancements in this domain are critical, as they can expand the applicability of NeRFs to practical scenarios where data is sparse, effectively bridging the gap between 2D and 3D data representation. Addressing the few-shot rendering challenge requires robust reconstruction techniques and a deep understanding of image data, given the inherently under-constrained nature of the problem.


Several works tackle the few-shot rendering problem~\cite{yu2021pixelnerf,chen2021mvsnerf,jain2021putting,wang2023sparsenerf,seo2023flipnerf,kim2022infonerf,yang2023freenerf,shi2024zerorf}, each with a unique approach but sharing the common goal of addressing the ambiguities of this ill-posed problem by introducing priors. Some methods incorporate \textit{data-driven priors}~\cite{yu2021pixelnerf,chen2021mvsnerf,jain2021putting,wang2023sparsenerf} by pre-training on diverse scenes~\cite{yu2021pixelnerf,chen2021mvsnerf} or leveraging robust pre-trained modules, such as vision-language~\cite{jain2021putting} or depth models~\cite{wang2023sparsenerf}.
A limitation of data-driven priors is that they often have difficulty generalizing their knowledge to new, unseen scenes.
For example, these priors work well with scenes similar to their training data on indoor objects but struggle with novel or diverse scenes on wild natural or in vivo scenes due to the huge domain gap between the scene's distribution.

Unlike data-driven priors, some approaches rely solely on the information available in the given training data and use \textit{explicit regularization}~\cite{seo2023flipnerf,kim2022infonerf,yang2023freenerf,shi2024zerorf}.
Our work follows this latter direction, imposing a prior that constrains the search space of learnable parameters, similar in spirit to FreeNeRF's use of frequency masking~\cite{yang2023freenerf} or ZeroRF's application of the Deep Image Prior~\cite{ulyanov2018deep,shi2024zerorf}.


Existing methods for few-shot rendering face significant limitations in practical scenarios. Approaches leveraging \textit{data-driven priors} are often computationally intensive, restricting their real-world applicability. Beyond the substantial computational cost of pre-training, most methods~\cite{yang2023freenerf,yu2021pixelnerf,chen2021mvsnerf,wang2023sparsenerf,jain2021putting} are built on the original NeRF~\cite{mildenhall2020nerf} or Mip-NeRF~\cite{barron2021mipnerf} frameworks, which are notoriously slow to train.
While various techniques exist to accelerate NeRF training~\cite{Chen2022ECCV,mueller2022instant,kerbl3Dgaussians}, only ZeroRF~\cite{shi2024zerorf}—to the best of our knowledge—applies accelerated representations, such as grid-based methods, to the \textit{few-shot setting without data priors}. ZeroRF adapts TensoRF~\cite{Chen2022ECCV} for this context but struggles to handle real-world scenes effectively, as acknowledged by its authors~\cite{shi2024zerorf}. Other approaches address few-shot rendering using 3D Gaussian splatting~\cite{li2024dngaussian,zhu2025fsgs}; however, these rely on depth information, placing them outside the scope of our comparisons.



In this work, we present \textbf{FourieRF} a method to parameterize the features of grid-based NeRF methods using a general prior based on the Fourier Transform. Our approach has virtually no computational overhead since we apply it per iteration. FourieRF is fast, robust, and effective across a range of scenes, from synthetic to real. To showcase our method we compare FourieRF against the best existing \textit{learning-free} methods
and show that FourieRF establishes a new state-of-the-art by delivering robust results in record time. In summary, our contributions are as follows:



\begin{itemize}
\item We demonstrate that by constraining the maximum Fourier frequency, it is possible to regulate the level of details in NeRF's scene representations in an artifact-free manner.
\item We show that the smooth shapes learned with low frequency accurately capture the scene's coarse geometry, providing robust and stable initialization in both few-shot and dense inputs.
\item We introduce a fast grid-based NeRF representation that band-limits feature grids’ frequency with a novel training curriculum. Our simple approach produces results on par with state-of-the-art while requiring a fraction of the time to converge.

\end{itemize}

\begin{figure}[t]
	\centering
	\includegraphics[width=\linewidth]{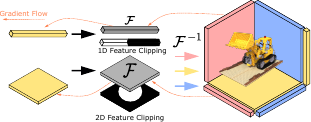}
	\caption{\textbf{Method illustration.} From left to right. Feature vectors and matrices are initialized in the spatial space. They are projected using the FFT. The Fourier coefficients are clipped using the masking procedure. Finally, the inverse FFT is applied to retrieve the smoothed features. }
    \vspace{-1em}
	\label{fig:method}
\end{figure}

%% file: sec/2_related_work.tex
\section{Related Work}
\label{sec:related_work}

\begin{figure*}[t]
	\centering
	\includegraphics[width=0.95\linewidth,trim={0 0 0 0},clip]{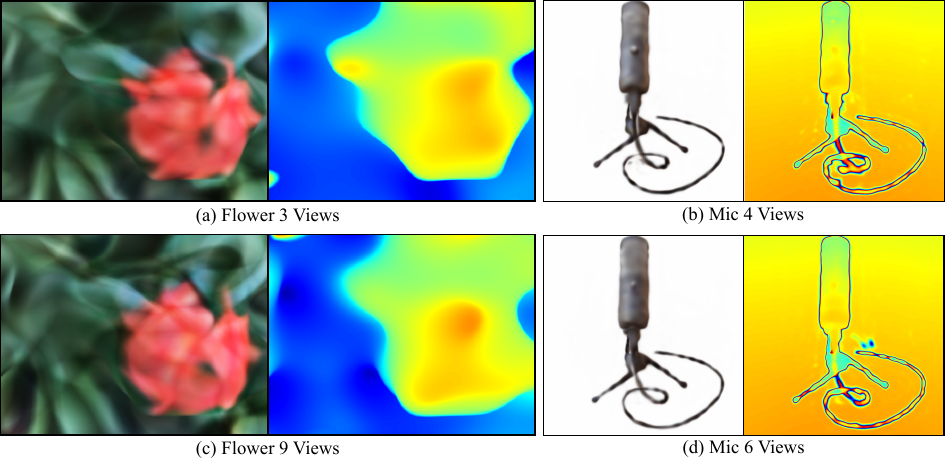}
	\caption{\textbf{Coarse Geometry Extraction.} Our method is capable of extracting correct coarse geometry from as little as 3 views. This coarse geometry remains relatively stable regardless of the number of views we input.}
    \vspace{-10px}
	\label{fig:coarse-geometry-extraction}
\end{figure*}

\paragraph{Radiance Field Representations}

The seminal work Neural Radiance Fields (NeRF) \cite{mildenhall2020nerf} uses deep neural architectures and a set of given images to produce photo-realistic novel-view synthesis (NVS). Subsequent grid methods, accelerate the pipeline by replacing the deep neural architecture with a 3D grid of features \cite{Chen2022ECCV}, or a multi-resolution hash grid \cite{mueller2022instant}. These grids' NeRF approaches were shown to converge orders of magnitudes faster than the original work while maintaining high-quality results. Encoding the information in an explicit 3D representation is a weak prior that leads to significantly easier learning. However, a common struggle of all NeRF representations is that they are \textit{data hungry}, they require several images to perform the NVS task properly. When given limited images, say 3 or 6, these models are extremely prone to overfitting; \textit{they produce incoherent geometry and images from novel points of view.}

\vspace{-1em}\paragraph{Few-shot Novel View Synthesis}

The problem described above is commonly referred to as the few-shot rendering problem. Approaches to address it can be categorized into two groups: methods that rely on data priors, such as scene depth~\cite{wang2023sparsenerf,zhu2025fsgs,li2024dngaussian} or diffusion models~\cite{wu2024reconfusion} to mitigate reconstruction ambiguities; and methods that operate without external data~\cite{yu2021pixelnerf,chen2021mvsnerf,jain2021putting,seo2023flipnerf,kim2022infonerf,yang2023freenerf,shi2024zerorf}. In this work, we focus on comparing our approach with state-of-the-art methods that do not rely on data priors, applying solely \textit{explicit regularization}, specifically FreeNeRF~\cite{yang2023freenerf} and ZeroRF~\cite{shi2024zerorf}.
We demonstrate that our method achieves state-of-the-art results while setting new records for speed, leveraging the Fourier transform to effectively control the complexity of features in a 3D grid.



\vspace{-1em}\paragraph{Compression Neural Fields}

The work FreeNeRF \cite{yang2023freenerf} shows there exists a link between the failure of few-shot neural rendering and the positional encoding used by deep neural network-based NeRF methods. Their method proposes to mask the encoding and progressively give it to the network. This allows for a coarse-to-fine reconstruction. Nevertheless, their take is specific to NeRF representations that are based on deep neural architecture, it is not trivial to apply it to accelerated grid methods. Therefore making FreeNeRF \textit{extremely} slow to train.

On the other hand, ZeroRF \cite{shi2024zerorf} leverages the Deep Image Prior \cite{ulyanov2018deep} to parameterize a grid NeRF representation. This work constitutes, to our knowledge, the first instance of an accelerated NeRF representation that specifically tackles the few-shot rendering problem. The authors, however, acknowledge the limited applicability of their method to \textit{real} (i.e., non-synthetic) scenes.

We present FourieRF a method to parameterize the features of grid-based NeRF representations. Our method trains in record time (approximately 10 minutes) and can tackle real and synthetic scenes.

%% file: sec/3_motivation.tex
\section{Motivation and Overview}
\label{sec:motivation}
\begin{figure*}[t]
	\centering
    \vspace{-1em}
	\includegraphics[width=\linewidth]{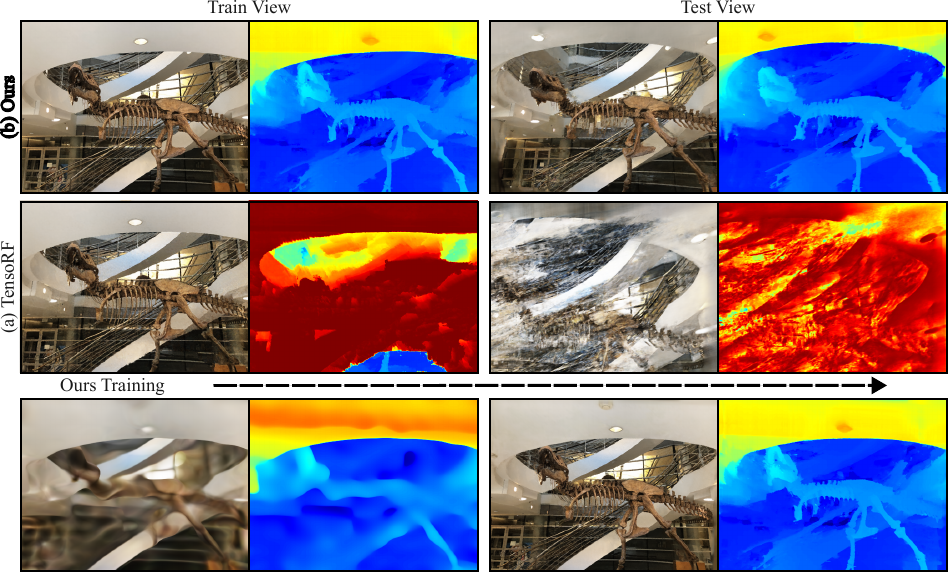}
	\caption{\textbf{Overfitting on the few-shot rendering problem.} ``Catastrophic overfitting" is a common behavior for standard NeRF representations on the few-shot rendering problem. Degenerate geometry is learned, which might result in plausible views near train inputs but does not generalize to novel views. }
    \vspace{-1em}
	\label{fig:fail_cases}
\end{figure*}

The few-shot rendering problem is inherently ill-posed.
With only a few images, there is insufficient information to uniquely determine the reconstructed 3D scene.
As a result, training a standard NeRF model on very limited data will inevitably lead to overfitting, where the model memorizes the few input images and captures noise instead of learning an underlying 3D structure.
When an overfitted NeRF tries to synthesize new views, the results will be inaccurate or unnatural, producing distorted images that are unrealistic and inconsistent with the true 3D structure of the scene~\cite{yang2023freenerf,yu2021pixelnerf}.


The problem highlighted by the FreeNeRF paper is that the vanilla NeRF's use of high-frequency positional encodings can cause the model to overfit drastically during the initial training iterations when given only a few input views. These high-frequency positional encoding inputs allow the model to quickly learn to fit the available data (the few training views) too well, but in doing so, it generates unrealistic or degenerate geometries, such as floaters. These geometries do not accurately represent the true structure of the scene but are instead arbitrarily constructed patches that help the network mimic patterns in input views.

This overfitting behavior is not limited to MLP-based NeRFs with positional encodings but is also observed in other NeRF representations. TensoRF~\cite{Chen2022ECCV} is an improved NeRF design that utilizes tensor decomposition to reduce computational complexity and memory usage while maintaining quality for 3D scene reconstruction. However, when directly used in few-shot scenarios, TensoRF also suffers from the overfitting problem.
In Fig.~\ref{fig:teaser}, we can see that TensoRF quickly fills the space with floaters that help the reconstruction on the set of limited views. These floaters do not generalize well to new views, but since they fit correctly the input views, removing them is challenging.



\begin{figure*}[t]
	\centering
    \vspace{-1em}
	\includegraphics[width=0.95\linewidth,trim={0 0 0 0},clip]{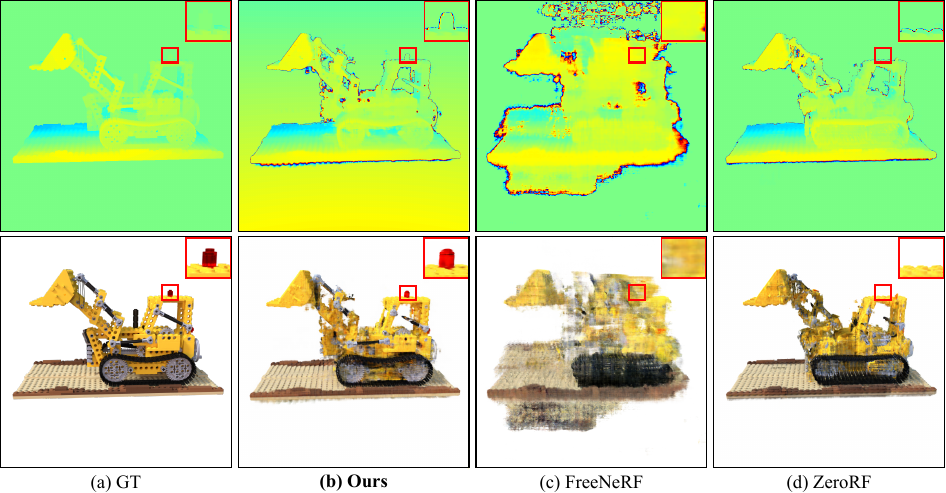}
	\caption{\textbf{Comparison on Blender Dataset.} In the Lego scene, trained with 4 views, we compare the performance of FreeNeRF, ZeroRF, and our method. ZeroRF renders a compact and clean reconstruction of the scene, however, at the cost of omitting some key details. FreeNeRF fails in this new setting due to its reliance on complex occlusion regularization. Despite employing a simple prior, FourieRF accurately captures both geometry and appearance, demonstrating a faithful reconstruction of the scene's details.}
	\label{fig:blender-qualitative}
\end{figure*}

FreeNeRF~\cite{yang2023freenerf} addresses overfitting by masking positional encodings for the MLP scene representation. This method is not directly applicable to grid-based methods like TensoRF~\cite{Chen2022ECCV}, limiting its application given the long training times of MLP-based NeRF. ZeroRF~\cite{shi2024zerorf} is the first work to use accelerated NeRF structures to tackle the few-shot rendering problem. It employs a convolutional network to generate feature maps of a TensoRF~\cite{Chen2022ECCV} representation, resulting in quickly trained ``clean" feature maps. However, this approach doesn't generalize well to real scenes~\cite{shi2024zerorf}, as its strong prior is not valid for non-synthetic scenes.

\textbf{FourieRF} deals with these two issues: it is an \textbf{accelerated} method, training in less than 10 minutes, that can tackle a wide variety of scenes, from \textbf{synthetic} to \textbf{real}. See Fig.~\ref{fig:method} for an illustration of our method. Using our approach, we can obtain correct coarse geometry from simple and complex scenes, see Fig.~\ref{fig:coarse-geometry-extraction}. 

Overall, our method is built on two key observations. First, we note that both strong overfitting and high-frequency artifacts typically occur early in the optimization process (see Fig.~\ref{fig:fail_cases}), and, if avoided in these early stages, they are significantly less prominent in the final result. Second, we note that by \textit{gradually} increasing the maximal Fourier frequency of the learned signal both significantly regularizes the learned NeRF, while at the same time, providing the network enough degrees of freedom to learn the fine details (in the final stages of the optimization). 

In other words, progressively increasing the available frequencies builds a robust trajectory to maintain the correctness of the shape as well as produce photo-realistic results \ref{fig:teaser}. This coarse to fine prior is not specific to any data type, and thus works in both real and synthetic scenes.

%% file: sec/4_method.tex
\section{Method}
\label{sec: method}

We apply our Fourier parameterization to representations introduced by TensoRF \cite{Chen2022ECCV}, increasing the maximal Fourier frequency in tensor decomposition gradually.


\subsection{Preliminaries}

The key idea behind grid-based NeRF representation is to represent the scene using a decomposed feature grid rather than a deep neural network. We denote the 3D tensor of features that represents the scene as $\mathcal{T}\in \mathbb{R}^{I\times J \times K}$. In our NeRF representation, we experiment with two different methods to decompose this 3D tensor:

\vspace{-1em}\paragraph{CANDECOMP/PARAFAC (CP) Decomposition}

\begin{figure*}[t]
    \vspace{-1em}
	\centering
	\includegraphics[width=0.95\linewidth]{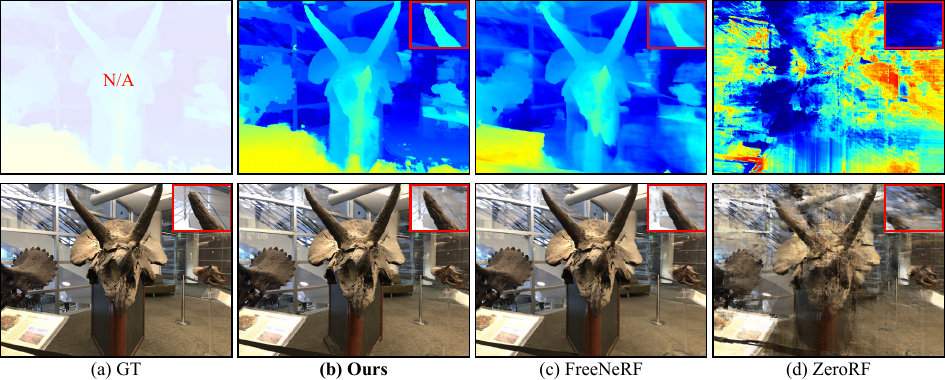}
	\caption{\textbf{Comparison in LLFF Dataset.} In the horns scene, we evaluated the performance of FreeNeRF, ZeroRF, and our method under a 3-view training setup. ZeroRF struggled to reconstruct coherent geometry, resulting in significant inconsistencies. FreeNeRF, while more stable, produced renders with notably blurred geometry, failing to capture fine details accurately. In contrast, FourieRF delivers sharper renders, faithfully reconstructing the \textit{key} geometric elements of the scene with high fidelity.}
    \vspace{-1em}
	\label{fig:llff_comparison}
\end{figure*}

In the CP decomposition, $\mathcal{T}$ is decomposed as a sum of outer products of vectors:
$$
\mathcal{T} = \sum_{r=1}^R v_r^1 \circ v_r^2 \circ v_r^3
$$
where $v_r^1 \circ v_r^2 \circ v_r^3$ corresponds to a rank-one tensor component, and $v_r^1 \in \mathbb{R}^I,v_r^2 \in \mathbb{R}^J,v_r^3 \in \mathbb{R}^K$ are factorized vectors of the three modes for the $r$-th component. 

CP factorization reduces space complexity from $\mathcal{O}(n^3)$ to $\mathcal{O}(n)$, and offers low-rank regularization at the same time in the optimization, making it a good candidate NeRF representation for few-shot reconstruction. On the other hand, CP sacrifices the rendering quality to minimize the rank of the decomposition.

\vspace{-1em}\paragraph{Vector-Matrix (VM) Decomposition}

Unlike CP factorization, VM decomposition enriches the product by using matrices. The decomposition is expressed as:
$$
\mathcal{T} = \sum_{r=1}^{R_1} v_r^1 \circ M_r^{2,3} + \sum_{r=1}^{R_2} v_r^2 \circ M_r^{1,3} + \sum_{r=1}^{R_3} v_r^3 \circ M_r^{1,2},
$$
where $M_r^{2,3} \in \mathcal{R}^{J\times K}, M_r^{1,3} \in \mathcal{R}^{I\times K}, M_r^{1,2} \in \mathcal{R}^{I\times J}$ are matrices for two of the three modes.

The VM decomposition reduces space complexity from $\mathcal{O}(n^3)$ to $\mathcal{O}(n^2)$. 
 For complex scenes, the VM decomposition reduces the number of components required to achieve the same expressivity as CP, leading to faster reconstruction and better rendering. Our method can be applied to any of these decompositions. In practice, applying our method to the VM decomposition allows us to model more complex effects, leading to better quantitative performance (See Table~\ref{tab:blender-results}).

\subsection{Fourier Parameterization}
\label{sec:fourier-parameterization}

As mentioned in Section \ref{sec:motivation}, the key observation behind our method is that objects' geometry and appearance can be learned in a coarse-to-fine manner based on their corresponding low to high frequencies in the underlying NeRF representation.
Our work makes the following claims:
(i) There are enough constraints in few-shot inputs to learn an accurate coarse geometry under low-frequency constraints;
(ii) Lower frequencies are easier to learn correctly than higher frequencies; 
(iii) Learning the next set of higher frequencies is more straightforward given a correct set of lower frequencies. 

Let us illustrate the above claims: in Fig.~\ref{fig:coarse-geometry-extraction} we can see that using our method we can establish a good base for scenes, even in the case of complex real scenarios. Moreover, Fig.~\ref{fig:teaser}, shows that given a good estimation of the low-frequencies of the scene, we can progressively add complexity to the object while maintaining a clean reconstruction.
In the following section, we demonstrate the process of parameterizing 1D and 2D features using our method. This parameterization allows us to begin with well-established coarse geometry and progressively incorporate fine details. In both cases, the principle remains the same: project the features into Fourier space and remove high frequencies up to a certain threshold. We make use of the discrete Fourier transform, so given a fixed grid size, there is a finite number of Fourier coefficients after transformation. The threshold we use is proportional to this finite number, please refer to the supplementary for typical values.

\begin{table*}[t]
	\centering
    \vspace{-1.5em}
	\caption{Quantitative comparison on Blender.}
	\label{tab:blender_comparison}
	\begin{tabular}{l l lcccccc}
		\toprule
		\multirow{2}{*}{Method} & \multirow{2}{*}{Prior} & \multicolumn{2}{c}{PSNR $\uparrow$} & \multicolumn{2}{c}{SSIM $\uparrow$} & \multicolumn{2}{c}{LPIPS $\downarrow$} \\
		\cmidrule(lr){3-4} \cmidrule(lr){5-6} \cmidrule(lr){7-8}
		& & 4 views & 6 views & 4 views & 6 views & 4 views & 6 views \\
		\midrule
		DietNeRF \cite{jain2021putting} & CLIP & 10.92 & 16.92 & 0.557 & 0.727 & 0.446 & 0.267 \\
		RegNeRF \cite{jain2021putting} & RealNVP & 9.93 & 9.82 & 0.419 & 0.685 & 0.572 & 0.580 \\
		\midrule
		TensoRF \cite{Chen2022ECCV}& No Prior & 18.656 & 21.652 & 0.798 & 0.844 & 0.216 & 0.165  \\
		\midrule
		ZeroRF \cite{shi2024zerorf}& Deep Network & \cellcolor{red!25} 21.94 & \cellcolor{red!25}24.73 & \cellcolor{orange!25} 0.856 & \cellcolor{red!25}0.889 & \cellcolor{red!25}0.139 & \cellcolor{red!25} 0.113 \\
		FreeNeRF \cite{yang2023freenerf} & Frequency & 18.81 & \cellcolor{yellow!25} 22.77 & \cellcolor{yellow!25} 0.808 & 0.865 & \cellcolor{yellow!25} 0.188 &   \cellcolor{yellow!25}0.1495 \\
		\midrule
		Ours - CP & Frequency & \cellcolor{yellow!25} 20.799 & 22.496 & \cellcolor{yellow!25}0.825 & \cellcolor{yellow!25}0.849 & 0.217 &  0.195  \\
		\textbf{Ours} & Frequency & \cellcolor{orange!25} 21.728 & \cellcolor{orange!25} 23.927 & \cellcolor{red!25} 0.858 & \cellcolor{orange!25}0.879 & \cellcolor{orange!25} 0.147 & \cellcolor{orange!25} 0.136  \\
		\bottomrule
	\end{tabular}
	\label{tab:blender-results}
\end{table*}

\begin{table*}[t]
	\centering
	\caption{Quantitative comparison on LLFF.}
	\label{tab:llff_comparison}
	\begin{tabular}{l lccccccccc}
		\toprule
		\multirow{2}{*}{Method} & \multicolumn{3}{c}{PSNR $\uparrow$} & \multicolumn{3}{c}{SSIM $\uparrow$} & \multicolumn{3}{c}{LPIPS $\downarrow$} \\
		\cmidrule(lr){2-4} \cmidrule(lr){5-7} \cmidrule(lr){8-10}
		& 3 views & 6 views & 9 views & 3 views & 6 views & 9 views & 3 views & 6 views & 9 views \\
		\midrule
		DietNeRF \cite{jain2021putting} & 14.94 & 21.75 & 24.28 & 0.370 & 0.717 & 0.801 & 0.496 & 0.248 & 0.183 \\
		RegNeRF \cite{niemeyer2022regnerf}& \cellcolor{yellow!25}19.08 & \cellcolor{yellow!25}23.10 & \cellcolor{yellow!25}24.86 & \cellcolor{yellow!25}0.587 & \cellcolor{yellow!25}0.760 &\cellcolor{yellow!25} 0.820 & \cellcolor{yellow!25} 0.336 &  \cellcolor{orange!25}0.206 &  \cellcolor{orange!25} 0.161 \\
		\midrule
		TensoRF \cite{Chen2022ECCV} & 14.292 & 18.183 & 23.677 & 0.315 & 0.576   & 0.777  & 0.545  & 0.370 & 0.213\\
		\midrule
		ZeroRF \cite{shi2024zerorf} & 16.74 & 21.371 & 22.425 & 0.434 &0.698 & 0.750 &  0.470 & 0.302 & 0.275 \\
		FreeNeRF \cite{yang2023freenerf}& \cellcolor{red!25} 19.63 & \cellcolor{red!25} 23.73 & \cellcolor{red!25} 25.13 &  \cellcolor{orange!25}0.612 &  \cellcolor{orange!25}0.779 &  \cellcolor{orange!25}0.827 &  \cellcolor{orange!25}0.308 & \cellcolor{red!25} 0.195 & \cellcolor{red!25}0.160 \\
		\midrule
		\textbf{Ours} &  \cellcolor{orange!25} 19.303 & \cellcolor{orange!25} 23.595 &  \cellcolor{orange!25}25.011 & \cellcolor{red!25} 0.636 & \cellcolor{red!25} 0.790 & \cellcolor{red!25} 0.830 & \cellcolor{red!25} 0.299 &  \cellcolor{yellow!25}0.210 &  \cellcolor{yellow!25}0.193  \\
		\bottomrule
	\end{tabular}
\end{table*}


\vspace{-1em}\paragraph{1D Features.}

For 1D features we have $\textbf{v} \in \mathcal{R}^d$, a $d$-dimensional feature. At time $t$, given a feature threshold $f_t$ we perform the following operations. The feature $\textbf{v}$ is projected into the Fourier space, then the Fourier coefficients are clipped using the threshold $f_t$; finally, we apply the inverse Fourier transform,
\begin{equation}
\hat{\textbf{v}} = \textbf{IFFT}(\textbf{FFT}(\textbf{v}) \odot \alpha(f_t))
\end{equation}
Call $d_f$ the dimension of $\textbf{FFT}(\textbf{v})$, then in practice the mask $\alpha(f_t)$ corresponds to an array of the same dimension where we keep cells up to index $t_\alpha(f_t) = d_f \times f_t$. Each cell $i$ is given by,
\begin{equation}
\alpha_i(f_t) = \begin{cases}
    1 \text{ if, } i <  t_\alpha(f_t)\\
    t_\alpha(f_t)- \floor*{t_\alpha(f_t)} \text{ if, } i = t_\alpha(f_t) \\
    0 \text{ otherwise} \\
\end{cases}
\end{equation}
The same 1D feature parameterization is applied to both, the CP and the VM decompositions.

\vspace{-1em}\paragraph{2D Features.}

For 2D features, we have $\textbf{w} \in \mathcal{R}^{d_1\times d_2}$ a matrix feature. We proceed as above,
\begin{equation}
\hat{\textbf{w}}= \textbf{IFFT}(\textbf{FFT}(\textbf{w}) \odot \beta(f_t))
\end{equation}
The difference mainly lies in the mask $\beta(f_t)$. In this case, we define a 2D mask with a circle centered at 
$c=\floor*{\frac{d_1}{2}},\floor*{\frac{d_2}{2}}$, of radius
$r=\frac{f_t}{2}\sqrt{2 \max(d_1,d_2)^2}$
This ensures that when $f_t$ reaches 100\% all parameters are used. The values outside of the circle are set to 0 to clip the corresponding coefficients. The 2D feature parameterization is only used in the VM decomposition.

\vspace{-1em}\paragraph{Progressive Inclusion of Coefficients.}
To smoothly control the frequency of our feature grid decomposition, we define a mask to clip the corresponding Fourier coefficients and progressively increase the frequency using a clipping threshold. 
The illustration in Fig.~\ref{fig:teaser} shows the progressive increase effects. 
When setting the clipping parameter, it is important to keep it sufficiently low initially to ensure the correct coarse geometry. Examples of successful choices of clipping can be seen in Fig.~\ref{fig:coarse-geometry-extraction}. 
In our method, we initially start with only $f_0=0.01\%$ of Fourier coefficients and then linearly increase the clipping parameter during training. To be specific, we update it every iteration as follows:
$
f_t = f_{t-1} + \Delta, \text{with } \Delta = \frac{1-f_0}{N}
$
Where $t$ is the iteration number, $f_0$ is the initial clipping, and $N$ is the number of iterations. The update is applied at the start of every iteration, before any gradient is accumulated, thus avoiding any differentiability problems.

\vspace{-1em}\paragraph{View Dependence}

In the few-shot learning setting, it is extremely challenging to learn view-dependent information. We hypothesize that the model can use directional information ($d$) and positional encodings to overfit to a limited set of views. In practice, we have found it sufficient to restrict the directional information provided to the model. We adopt a similar approach to ZeroRF~\cite{shi2024zerorf} by using a simplified decoder that \textit{does not} use positional encodings for features or view directions.


%% file: sec/5_experiments.tex
\section{Experiments}

\subsection{Setup}
\paragraph{Datasets \& metrics.}

Our method \textbf{FourieRF} can process a wide variety of scenes. We thus test it on synthetic and real scenes. The \textbf{NeRF-synthetic} dataset~\cite{mildenhall2020nerf} was rendered using Blender containing 8 objects with complex material and geometric information. We use ZeroRF's~\cite{shi2024zerorf} \textit{same setting to train and evaluate} our method (with the number of views ranging from 4 to 6). The \textbf{LLFF}~\cite{mildenhall2019local} contains 8 real scenes. We use RegNeRF's~\cite{niemeyer2022regnerf} \textit{same setting to train and evaluate} our method (training on 3, 6, or 9 views).

\vspace{-1em}\paragraph{Implementation.}

FourieRF can be easily added to the TensoRF~\cite{Chen2022ECCV}. Our method can enhance both the performance of the CP and VM decomposition, and in each case we create a class that inherits from the respective TensoRF decomposition. We find in practice that the VM decomposition leads to better results (See Table.~\ref{tab:blender-results}), so we use it in all the experiments unless otherwise stated. Please see the supplementary material for more information about our code and the hyper parameters used.

\vspace{-1em}\paragraph{Baselines}
Our first baseline is vanilla TensoRF-VM~\cite{Chen2022ECCV}, as it is the foundation upon which our method is built. We also compare our method to ZeroRF~\cite{shi2024zerorf} and FreeNeRF~\cite{yang2023freenerf}, the most recent and directly comparable baselines. ZeroRF specializes in the reconstruction of synthetic scenes and is an accelerated method that trains in around 30 minutes, but, as noted earlier, it struggles with real scenes. In contrast, FreeNeRF can process any type of scene~\cite{yang2023freenerf}, but its training time is extremely long (approximately one day). For completeness, we also include quantitative comparisons with well-established baselines such as DietNeRF~\cite{jain2021putting}, which leverages data priors, and RegNeRF~\cite{niemeyer2022regnerf}, which relies on geometrical regularizations.

\subsection{Results}

Our method achieves performance that is on par with the state-of-the-art (SOTA) approaches, while being significantly faster than comparable methods. We refer the reader to the supplementary material, where we showcase a video presentation and animated results.

Indeed in Table~\ref{tab:execution_time_comparison}, we can see that our method's training time is even faster than TensoRF. Our Fourier parameterization is done per iteration at virtually no cost. Moreover, as seen in Fig.~\ref{fig:fail_cases}, TensoRF's scene representation is filled with floaters. This hinders training, filling the scene with noise, thus slowing down the whole procedure. When compared to the other accelerated method, ZeroRF \cite{shi2024zerorf}, we see that their Deep Image Prior \cite{ulyanov2018deep}, comes at the cost of evaluating an expensive convolutional neural network. Our method is by far the fastest to converge in the few-shot rendering task.

The results of our quantitative evaluation are showcased in Table~\ref{tab:blender-results} and Table~\ref{tab:llff_comparison}. 
In the synthetic dataset, we greatly outperform all MLP-based methods by a significant margin. We achieve this while training an order of magnitude faster, and without the use, of the hard-to-tune, occlusion regularization used by FreeNeRF \cite{yang2023freenerf}. The SOTA in this dataset is the accelerated method ZeroRF \cite{shi2024zerorf}. However, we achieve similar results while training over 5 times faster, and Fig.~\ref{fig:blender-qualitative} shows that ZeroRF's prior can lead to the omission of key geometry. Despite its state-of-the-art performance on synthetic datasets, ZeroRF~\cite{shi2024zerorf} fails to handle real scenes effectively. This goes to show, that their Deep Image Prior \cite{ulyanov2018deep} does not generalize to diverse scenes. Our method achieves results that are on par with FreeNeRF \cite{yang2023freenerf}, while training \textit{over 30 times faster.} Moreover, in Fig.~\ref{fig:llff_comparison}, we see that the shapes we extract are in some cases smoother than FreeNeRF's. 

These results demonstrate that we have introduced an exceptionally \textit{simple} and \textit{flexible} baseline. Our method performs well across a variety of scenes while training at record speeds, highlighting its practicality and ease of use.

\vspace{-1em}\paragraph{Ablations.}

\begin{figure}[t]
	\centering
    \vspace{-1em} 
	\includegraphics[width=\linewidth,trim={0 0 0 0},clip]{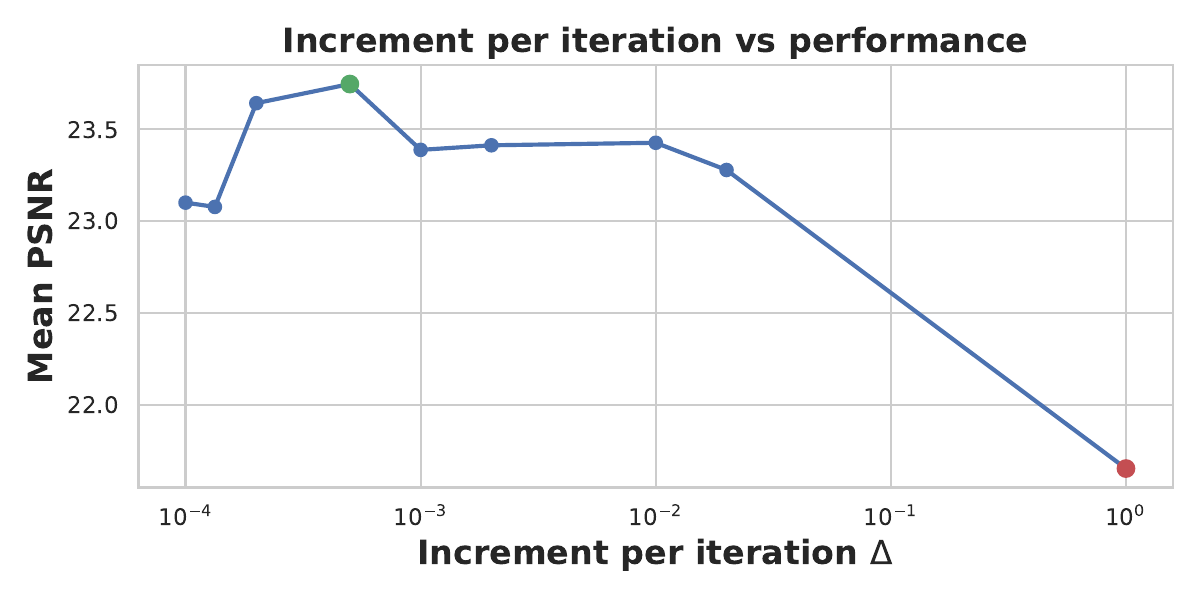}
    \vspace{-2em} %
	\caption{\textbf{Choice of $\Delta$ vs performance.} We investigated the effect of varying the speed at which high-frequencies are integrated during training, using the Blender Dataset with 6 views. The baseline performance without our method is highlighted in red, while our best result is shown in green.}
	\label{fig:delta-vs-performance}
\end{figure}

The progressive inclusion of complexity is a key aspect of our method. In Fig.~\ref{fig:teaser}, we illustrate a successful training trajectory. However, determining ``how quickly" this complexity is integrated—i.e., setting the parameter $\Delta$ defined in Section~\ref{sec:fourier-parameterization}—is crucial. Fig.~\ref{fig:delta-vs-performance} shows the relationship between the choice of this parameter and the PSNR obtained when training and testing across all scenes from the Blender synthetic dataset (both using 6 views). Fig.~\ref{fig:delta-vs-performance} shows that for large $\Delta$, the method converges to the baseline, TensoRF. In practice, choosing a sufficiently small increment should yield adequate performance. We observe a slight decrease in performance for smaller values of $\Delta$, which is likely due to the fixed 10k training iterations used in the experiment. For the smallest increments, the model simply did not have enough training time.

\begin{table}[t]
	\centering
    \vspace{-0.5em}
	\caption{Global execution time comparison relative to TensoRF. Training on the Blender Dataset for 10k iterations.}
	\label{tab:execution_time_comparison}
	\begin{tabular}{lc}
		\toprule
		Method & Training Time \\
		\midrule
		TensoRF \cite{Chen2022ECCV} & \cellcolor{orange!25}{$1.0 \times$}\\
		ZeroRF \cite{shi2024zerorf} &  \cellcolor{yellow!25}{$5.181 \times$}\\ 
		FreeNeRF \cite{yang2023freenerf} &  $35.71 \times$ \\
		\textbf{Ours} & \cellcolor{red!25}{$0.93 \times$} \\
		\bottomrule
	\end{tabular}
    \vspace{-1em}
\end{table}

Finally, our method stands as the \textit{fastest} baseline available for the few-shot rendering problem. As shown in Table~\ref{tab:execution_time_comparison}, our execution time is virtually identical to that of vanilla TensoRF~\cite{Chen2022ECCV}, with our method being slightly faster because we do not predict ``useless" floaters. Additionally, we are more than five times faster than ZeroRF~\cite{shi2024zerorf}, the only other accelerated method addressing the few-shot rendering problem.




%% file: sec/6_conclusion.tex
\section{Conclusion \& Future Work}

In this work, we introduce \textbf{FourieRF}, a novel approach for achieving fast and high-quality reconstruction in the few-shot setting. Our method effectively parameterizes features through an explicit curriculum training procedure, incrementally increasing scene complexity during optimization. Experimental results show that the prior induced by our approach is both robust and adaptable across a wide variety of scenes, establishing \textbf{FourieRF} as a strong and versatile baseline for the few-shot rendering problem. While our approach significantly reduces artifacts, it may still lead to reconstruction errors in severely under-constrained scenarios, particularly where view occlusion leaves parts of the shape uncovered.  In the future, our method could be enhanced by integrating foundation models to complete missing parts using large data-driven priors.

\paragraph{Acknowledgement.} Parts of this work were supported by the ERC Consolidator Grant 101087347 (VEGA) and the ANR AI Chair AIGRETTE.


%% file: sec/X_suppl.tex
\clearpage
\setcounter{page}{1}
\maketitlesupplementary

\section*{A. Supplementary Results on NeRF Synthetic and LLFF dataset}

First, we note that as part of the supplementary materials we have \textbf{included a interactive website}, which provides qualitative comparisons with two closest baselines to our method (methods that have comparable running times): TensoRF \cite{Chen2022ECCV} and ZeroRF \cite{shi2024zerorf}. As can be seen in the provided website, our method significantly outperforms those baselines in terms of quality of reconstructions in the few-shot setting, and has fewer artefacts, especially when considering very few input views. We provide qualitative comparisons both in terms of the novel view synthesis (RGB) as well as depth estimation compared to these baseline approaches. We encourage the reader to consider the videos provided in the interactive website (please allow a few seconds to load the videos) to see the improvement provided by our method.

In addition, below we include details on the statistics of our evaluations on the LLFF dataset in tables \ref{tab:llff-details-3},\ref{tab:llff-details-6},\ref{tab:llff-details-9} and on the NeRF synthetic dataset in tables \ref{tab:nerf_synthetic_4_details} and \ref{tab:nerf_synthetic_6_details}. For the LLFF dataset we reproduced the ZeroRF experiments to obtain the per scene score.

\begin{table*}[ht]
    \centering
    \caption{Details quantitative comparison on the LLFF real dataset 3 views.}
    \begin{tabular}{lcccccccccc}
        \toprule
        Method & Statistic & fortress & room & horns & orchids & leaves & fern & flower & trex & mean\\
        \midrule
        \multirow{3}{*}{FreeNeRF~\cite{yang2023freenerf}}
          & PSNR $\uparrow$  & \first 23.437 & \first 22.020 & \first 18.506 & \first 15.286 & \second 16.250 & \first 21.187 & \second 20.413 & \second 19.941 & \first 19.630	  \\
        & SSIM $\uparrow$  & \first 0.583 & \first 0.834 & \second 0.585 & \second 0.407 & \second 0.521 & \second 0.662 & \second 0.617 & \second 0.687 & \second 0.612	 \\
        & LPIPS $\downarrow$ & \second 0.319 & \first 0.190 & \second 0.355 & \second 0.377 & \second 0.350 & \second 0.286 & \second 0.291 & \second 0.297 & \second 0.308   \\

        \midrule
        \multirow{3}{*}{ZeroRF~\cite{shi2024zerorf}}
          & PSNR $\uparrow$  & \third 20.633 & \third 18.833 & \third 13.688 & \third 13.900 & \third 16.275 & \third 18.700 & \third 17.880 & \third 16.786 & \third 17.087  \\
        & SSIM $\uparrow$  & \third 0.435 & \third 0.663 & \third 0.233 & \third 0.275 & \third 0.533 & \third 0.523 & \third 0.490 & \third 0.517 & \third 0.459  \\
        & LPIPS $\downarrow$ & \third 0.386 & \third 0.392 & \third 0.612 & \third 0.527 & \third 0.398 & \third 0.422 & \third 0.423 & \third 0.451 & \third 0.451  \\
        \midrule
        \multirow{3}{*}{Ours}
        & PSNR $\uparrow$  & \second 22.109 & \second 20.271 & \second 18.290 & \second 15.103 & \first 16.524 & \second 20.965 & \first 21.062 & \first 20.103 & \second 19.303  \\
        & SSIM $\uparrow$  & \second 0.573 & \second 0.792 & \first 0.627 & \first 0.422 & \first 0.587 & \first 0.667 & \first 0.674 & \first 0.745 & \first 0.636  \\
        & LPIPS $\downarrow$ & \first 0.305 & \second 0.294 & \first 0.336 & \first 0.359 & \first 0.290 & \first 0.271 & \first 0.266 & \first 0.271 & \first 0.299  \\
        \bottomrule
    \end{tabular}
    \label{tab:llff-details-3}
\end{table*}  

\begin{table*}[ht]
    \centering
    \caption{Details quantitative comparison on the LLFF real dataset 6 views.}
    \begin{tabular}{lcccccccccc}
        \toprule
        Method & Statistic & fortress & room & horns & orchids & leaves & fern & flower& trex & mean\\
        \midrule
        \multirow{3}{*}{FreeNeRF~\cite{yang2023freenerf}}
          & PSNR $\uparrow$  & \second28.728 & \second27.302 &  \first23.592 & \second17.263 & \second19.047 &  \first24.647 &  \first24.665 &  \first24.596 &  \first23.730	  \\
        & SSIM  $\uparrow$  & \second0.832 & \second0.910 & \second0.792 & \second0.555 & \second0.685 & \second0.796 & \second0.797 &  \first0.864 & \second0.779	 \\
        & LPIPS $\downarrow$ & \second0.162 &  \first0.117 &  \first0.218 &  \first0.291 & \second0.260 &  \first0.196 &  \first0.162 &  \first0.154 & \first0.195  \\
          
        \midrule
        \multirow{3}{*}{ZeroRF~\cite{shi2024zerorf}}
           & PSNR $\uparrow$  & \third 23.767 & \third 27.083 & \third 19.188 & \third 14.425 & \third 18.475 & \third 23.533 & \third 21.780 & \third 21.957 & \third 21.276  \\
        & SSIM  $\uparrow$  & \third 0.802 & \third 0.880 & \third 0.606 & \third 0.318 & \third 0.670 & \third 0.753 & \third 0.712 & \third 0.796 & \third 0.692  \\
        & LPIPS $\downarrow$ & \third 0.195 & \third 0.211 & \third 0.387 & \third 0.519 & \third 0.319 & \third 0.280 & \third 0.277 & \third 0.279 & \third 0.308  \\
        \midrule
        \multirow{3}{*}{Ours}
        & PSNR $\uparrow$  &  \first29.031 &  \first28.792 & \second23.273 &  \first17.484 &  \first19.187 & \second24.466 & \second24.510 & \second22.019 & \second23.595  \\
        & SSIM $\uparrow$  &  \first0.878 &  \first0.920 &  \first0.815 &  \first0.558 &  \first0.727 &  \first0.792 &  \first0.822 & \second0.810 &  \first0.790  \\
        & LPIPS $\downarrow$ &  \first0.144 & \second0.165 & \second0.217 & \second0.313 &  \first0.214 & \second0.210 & \second 0.174 & \second0.243 & \second0.210  \\
        \bottomrule
    \end{tabular}
    \label{tab:llff-details-6}
\end{table*}  

\begin{table*}[ht]
    \centering
    \caption{Details quantitative comparison on the LLFF real dataset 9 views.}
    \begin{tabular}{lcccccccccc}
        \toprule
        Method & Statistic & fortress & room & horns & orchids & leaves & fern & flower & trex & mean\\
        \midrule
        \multirow{3}{*}{FreeNeRF~\cite{yang2023freenerf}} 
         & PSNR $\uparrow$ & \second 29.421 & \first 29.927 & \first 25.154 & \first 19.083 & \second 20.678 & \first 26.073 & \second 26.182 & \second 24.522 & \first 25.130 \\
        & SSIM $\uparrow$ & \second 0.865 & \first0.938 & \second 0.846 & \first 0.662 & \second 0.756 & \first 0.831 & \second 0.843 & \second 0.875 & \second 0.827	 \\
        & LPIPS $\downarrow$ & \first 0.124 & \first 0.091 & \first 0.174 & \first 0.237 & \second 0.222 & \first 0.159 & \first 0.133 & \first 0.139 & \first 0.16  \\

        \midrule
        \multirow{3}{*}{ZeroRF~\cite{shi2024zerorf}}
           & PSNR $\uparrow$ & \third 24.350 & \third26.883 & \third21.675 & \third16.125 & \third19.200 & \third24.400 & \third23.240 & \third24.629 & \third22.563  \\
        & SSIM $\uparrow$ &    \third0.797 & \third0.903 & \third0.733 & \third0.465 & \third0.700 & \third0.787 & \third0.762 & \third0.850 & \third0.750  \\
        & LPIPS $\downarrow$ & \third0.195 & \third0.189 & \third0.314 & \third0.424 & \third0.300 & \third0.242 & \third0.250 & \third0.229 &\third 0.268  \\
        \midrule
        \multirow{3}{*}{Ours}
        & PSNR $\uparrow$ &    \first 29.567 & \second 29.011 & \second 24.799 & \second 19.046 & \first 20.839 & \second 25.774 & \first  26.488 & \first 24.562 & \second 25.011  \\
        & SSIM $\uparrow$ &    \first 0.881 & \second 0.931 & \first 0.860 & \second 0.636 & \first 0.775 & \second 0.825 & \first 0.854 & \first 0.876 & \first 0.830  \\
        & LPIPS $\downarrow$ & \second 0.153 & \second 0.171 & \second 0.194 & \second 0.283 & \first 0.200 & \second 0.187 & \second 0.158 & \second 0.198 & \second 0.193  \\
        \bottomrule
    \end{tabular}
    \label{tab:llff-details-9}
\end{table*}  

\begin{table*}[ht]
    \centering
    \caption{Details quantitative comparison on the NeRF synthetic dataset 4 views.}
    \begin{tabular}{lcccccccccc}
        \toprule
        Method & Statistic & chair & drums & ficus & hotdog & lego & materials & mic & ship & mean\\
        \midrule
        \multirow{3}{*}{FreeNeRF~\cite{yang2023freenerf}}
          & PSNR $\uparrow$ & \third20.22 & \third14.99 & \third17.35 & \third23.58 & \third20.43 & \second 21.36 & \third15.05 & \third 17.52 & \cellcolor{yellow!25}{18.81} \\
          & SSIM $\uparrow$ & \third0.843 & \third0.746 & \third0.809 & \third0.899 & \third0.818 & \second 0.857 & \third0.802 & \third0.687 & \cellcolor{yellow!25}{0.808} \\
          & LPIPS $\downarrow$ & \third0.109 & \third0.280 & \third0.144 & \third0.108 & \third0.156 & \third0.174& \third0.218& \third0.318 & \cellcolor{yellow!25}{0.188} \\
        \midrule
        \multirow{3}{*}{ZeroRF~\cite{shi2024zerorf}}
          & PSNR $\uparrow$ & \second 23.04 & \second 16.91 & \first 20.12 & \first 29.11 & \second 22.11 & \third20.50 & \first 24.76 & \second 19.01 & \cellcolor{red!25}{21.94}\\
          & SSIM $\uparrow$ & \second 0.880 & \second 0.791 & \first 0.866 & \first 0.944 & \second 0.868 & \third0.848 & \first 0.944 & \second 0.707 & \cellcolor{orange!25}{0.856}\\
          & LPIPS $\downarrow$ &\first 0.074 & \first 0.131 & \first 0.100 & \first 0.075 & \first 0.085 &  \second 0.132 & \first 0.050 & \first 0.256 & \cellcolor{red!25}{0.113}\\
        \midrule
        \multirow{3}{*}{Ours}
        & PSNR $\uparrow$ & \first 24.13 & \first 17.33 & \second 18.56 & \second27.26 & \first 22.41 & \first 21.15 & \second23.35 & \first 19.64 & \cellcolor{orange!25}{21.73}  \\
         & SSIM $\uparrow$ & \first 0.895 & \first 0.804 & \second 0.848 & \second 0.933 & \first 0.871 & \first 0.858 & \second0.929 & \first 0.724 & \cellcolor{red!25}{0.858}  \\
         & LPIPS $\downarrow$ & \second 0.107 & \second 0.206 & \second 0.120 &  \second 0.088 & \second 0.122 & \first 0.129 & \second 0.056 & \second0.283 & \cellcolor{orange!25}{0.139}  \\
        \bottomrule
    \end{tabular}
    \label{tab:nerf_synthetic_4_details}
\end{table*}

\begin{table*}[ht]
    \centering
    \caption{Details quantitative comparison on the NeRF synthetic dataset 6 views.}
    \begin{tabular}{lcccccccccc}
        \toprule
        Method & Statistic & chair & drums & ficus & hotdog & lego & materials & mic & ship & mean\\
        \midrule
        \multirow{3}{*}{FreeNeRF~\cite{yang2023freenerf}}
          & PSNR $\uparrow$ & \second26.72 & \third18.16 & \third18.46 & \third27.18 & \third24.32 & \first 21.63 & \third25.64 & \third20.23 & \third22.77 \\
          & SSIM $\uparrow$ & \third0.916 & \third0.827 & \third0.840 & \third0.929 & \third0.887 & \second0.853 & \second0.942 & \third0.729 & \third0.865 \\
          & LPIPS $\downarrow$ & \second0.071 & \second0.176 & \second0.161 & \second0.096 & \third0.132 & \third0.202 & \second0.066 & \third0.290 & \third0.149 \\
        \midrule
        \multirow{3}{*}{ZeroRF~\cite{shi2024zerorf}}
          & PSNR $\uparrow$ & \first 27.62 & \first 20.88 & \first 22.21 & \first29.93 & \second26.26 & \third21.41 & \first27.40 & \second22.13 & \first 24.73 \\
          & SSIM $\uparrow$ & \first 0.926 & \first 0.869 & \first 0.898 & \first 0.949 & \second0.913 & \third0.849 & \first0.954 & \second0.756 & \first 0.889\\
          & LPIPS &\first 0.074 & \first 0.131 & \first 0.100 & \first 0.075 & \first 0.085 & \first 0.132 &\first 0.050 & \first 0.256 & \first 0.113\\
        \midrule
        \multirow{3}{*}{Ours}
         & PSNR $\uparrow$ & \third26.62 & \second19.30 & \second19.43 & \second28.84 & \first 27.09 & \second21.46 & \second25.78 & \first 22.89 & \second23.93  \\
         & SSIM $\uparrow$ & \second0.918 & \second0.838 & \second0.860 & \second0.939 & \first 0.915 & \first 0.856 & \second0.942 & \first 0.767 & \second0.879  \\
         & LPIPS $\downarrow$ & \third0.095 & \third0.182 & \third0.124 & \third0.108 &\second0.103 & \second0.141 & \third0.072 & \second0.261 & \second0.136  \\
        \bottomrule
    \end{tabular}
    \label{tab:nerf_synthetic_6_details}
\end{table*}

\section*{B. Details on Implementation and settings}

Finally, we provide details surrounding the settings of our implementation and experiments. All of our experiments were run in an nvidia RTX 4090 graphics card.
We build our code base in top of the TensoRF \cite{Chen2022ECCV} repository. Our repository can be found in the following link: \url{https://github.com/diego1401/FourieRF}.

Our method uses the AdamW optimizer \cite{kingma2014adam,loshchilov2017decoupled} with $\beta_1  = 0.9, \beta_2= 0.98$, and a weight decay of $0.2$ for synthetic scenes and $0$ for real scenes. When performing frequency-control on real scenes we have to deal with matter appearing in front of the camera as a form of overfitting, as highlighted by FreeNeRF \cite{yang2023freenerf}. We find that applying their occlusion regularization works without any modification in our pipeline, thus we use their hyper parameters to compute our metrics. Moreover, we note that it is also efficient to use the gradient scaling introduced in the Floaters No More paper \cite{philip2023floaters}, this approach does not require to set a hyper parameter. We train for 10k iterations to match with our baselines, mainly ZeroRF \cite{shi2024zerorf}.

The key hyper parameters of our method differ in the synthetic and real datasets. 
This can be attributed to the fact that the synthetic dataset has a solid color, white background, which alters the behavior of our method.

For the synthetic dataset, the clipping threshold is initialized as $f_0=0.3$, and it is linearly increased with $\delta = \frac{1}{2 000}= 2\times10^{-3}$. We use the same configuration parameters as TensoRF \cite{Chen2022ECCV} with the following differences. We apply a TV loss (with weight $w_{TV} = 1.0$) on the appearance and density features. We find that setting the weight decay to $0.2$ in the optimizer is the key to removing floaters (in our method and in ZeroRF \cite{shi2024zerorf}).

For the real dataset, the clipping threshold is initialized as $f_0=0.01$, and it is linearly increased until the end of training, i.e. $\delta = \frac{1}{10 000}= 10^{-4}$. We use the same configuration parameters as TensoRF \cite{Chen2022ECCV} with the following differences. We apply a TV loss (with weight $w_{TV} = 1.0$) on the appearance and density features, and an L1 loss (with weight $w_{L1} = 10^{-4}$) on the density features. We find that applying the L1 loss in this type of scenes is more efficient than setting a weight decay for the optimizer.